\documentclass[letterpaper, 10 pt, conference]{ieeeconf}  

\IEEEoverridecommandlockouts                              

\UseRawInputEncoding

\usepackage{graphics} 
\usepackage{epsfig} 
\usepackage{mathptmx} 
\usepackage{amsmath} 
\usepackage{amssymb}  
\usepackage[ruled,linesnumbered]{algorithm2e}
\usepackage{algorithmic}
\usepackage{color}
\usepackage{multirow}
\usepackage{upgreek}
\usepackage[T1]{fontenc}

\title{\LARGE \bf
	Automated Sperm Morphology Analysis Based on Instance-Aware Part Segmentation
}

\author{Wenyuan Chen$^{1}$$^{\dagger}$, Haocong Song$^{1}$$^{\dagger}$, Changsheng Dai$^{2}$, Aojun Jiang$^{1}$,  Guanqiao Shan$^{1}$, \\ Hang Liu$^{1}$, Yanlong Zhou$^{3}$, Khaled Abdalla$^{4}$, Shivani N Dhanani$^{4}$, Katy Fatemeh Moosavi$^{4}$, Shruti Pathak$^{4}$, \\ Clifford Librach$^{4}$, Zhuoran Zhang$^{5}$, and Yu Sun$^{1}$
\thanks{$^{1}$Robotics Institute, University of Toronto, Canada. $^{2}$School of Mechanical Engineering, Dalian University of Technology, Dalian, China. $^{3}$School Of Artificial Intelligence, Henan University, Henan, China. $^{4}$CReATe Fertility Center, Toronto, Canada. $^{5}$School of Science and Engineering, The Chinese University of Hong Kong, Shenzhen, China. }%
\thanks{$^{\dagger}$ The first two authors contribute equally to this work.}%
\thanks{\textit{Corresponding authors: Changsheng Dai (daichangsheng@dlut.edu.cn); Zhuoran Zhang (zhangzhuoran@cuhk.edu.cn); Clifford Librach (drlbrach@createivf.com); Yu Sun (yu.sun@utoronto.ca).}}%
}

\begin{document}

	\maketitle
	\thispagestyle{empty}
	\pagestyle{empty}

	\begin{abstract}
		
		Traditional sperm morphology analysis is based on tedious manual annotation. Automated morphology analysis of a high number of sperm requires accurate segmentation of each sperm part and quantitative morphology evaluation. State-of-the-art instance-aware part segmentation networks follow a ``detect-then-segment" paradigm. However, due to sperm's slim shape, their segmentation suffers from large context loss and feature distortion due to  bounding box cropping and resizing during ROI Align. Moreover, morphology measurement of sperm tail is demanding because of the long and curved shape and its uneven width. This paper presents automated techniques to measure sperm morphology parameters automatically and quantitatively. A novel attention-based instance-aware part segmentation network is designed to reconstruct lost contexts outside bounding boxes and to fix distorted features, by refining preliminary segmented masks through merging features extracted by feature pyramid network. An automated centerline-based tail morphology measurement method is also proposed, in which an outlier filtering method and endpoint detection algorithm are designed to accurately reconstruct tail endpoints. Experimental results demonstrate that the proposed network outperformed the state-of-the-art top-down RP-R-CNN by $9.2\%$ AP$^{p}_{vol}$, and the proposed automated tail morphology measurement method achieved high measurement accuracies of $95.34\%, 96.39\%, 91.20\%$ for length, width and curvature, respectively.
		
	\end{abstract}

	\section{INTRODUCTION}
	
	Quantitative analysis of sperm morphology is vital for diagnosing male infertility~\cite{murray2012effect}. Traditionally, sperm morphology analysis is performed manually by clinical staff who empirically follow the World Health Organization (WHO) guidelines~\cite{WHO} to analyze each sperm. Specifically, an experienced operator carefully observes zoomed-in sperm images to performs pixel-by-pixel annotation to label 5 contours for each part of the sperm (i.e., acrosome, vacuole, nucleus, midpiece and tail). Based on the annotated parts/contours, sperm morphology parameters are then manually calculated. Manual analysis of sperm morphology is a highly tedious process. As required by WHO, the morphology of at least 200 sperm should be analyzed within each semen sample, leading to manual annotation of $>1,000$ contours per patient sample, which calls for techniques to automate the analysis task.
	
\begin{figure}[!t]
		\centering
		\includegraphics[width=3.0in]{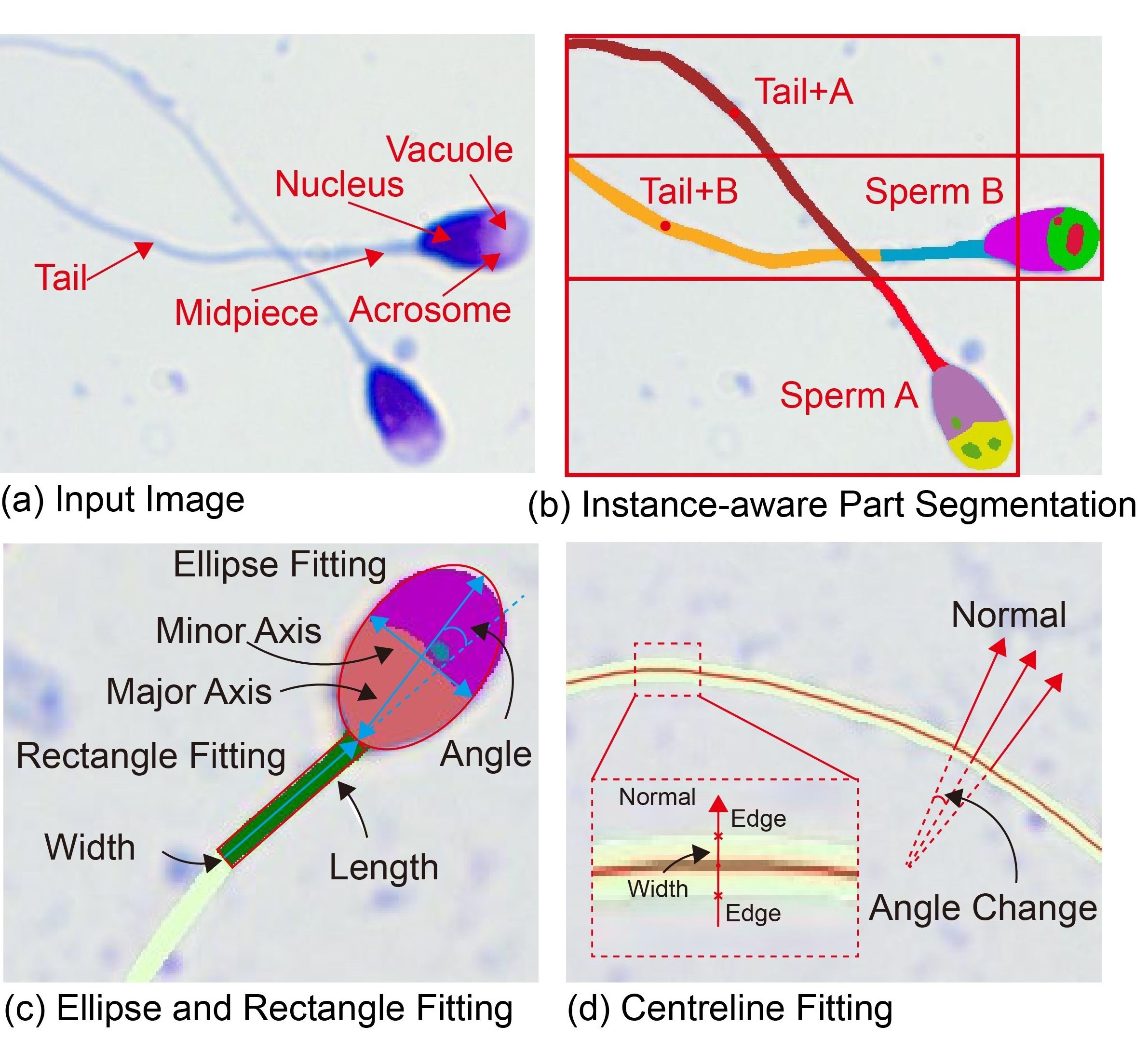}
		\caption{(a) An image of stained sperm. Each sperm is to be segmented into five parts: acrosome, vacuole, nucleus, midpiece and tail. (b) Instance-aware part segmentation that not only distinguishes different sperm, but also segments parts for each sperm. (c) Ellipse and rectangle fitting for measuring sperm head and midpiece morphology parameters. (d) Centerline fitting for measuring sperm tail morphology parameters.}
		\label{fig1}
	\end{figure}

	To achieve automated sperm morphology measurement, firstly, an instance-level sperm parsing is needed that not only correctly detects sperm but also accurately segments the parts for each sperm (see Fig.~\ref{fig1}(a)\&(b)). Secondly, an automated morphology measurement method is needed for quantitatively calculating morphology parameters for each sperm part based on the segmented mask.
	
\begin{figure}[!t]
		\centering
		\includegraphics[width=2.8in]{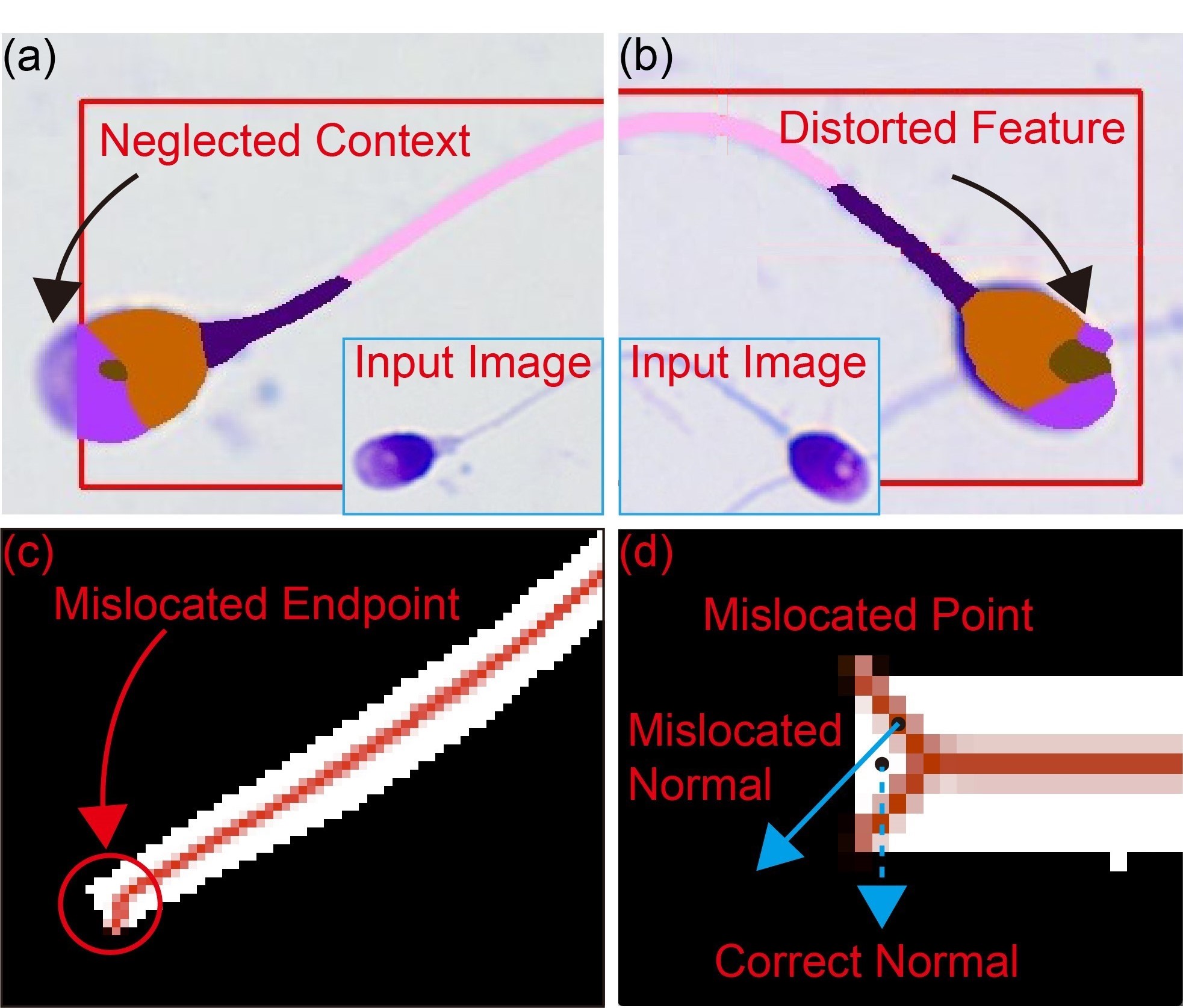}
		\caption{(a) Context loss due to bounding box cropping. (b) Feature distortion due to resizing in ROI Align. (c) Endpoints mislocated in Steger-based methods. (d) The normal of endpoints is mislocated due to the influence of gradient from the intersecting edge.}
		\label{fig2}
	\end{figure}

	For the first step of instance-aware part segmentation (see Fig.~\ref{fig1}(b)), state-of-the-art instance-aware part segmentation networks from the human parsing field employs a ``detect-then-segment" paradigm~\cite{yang2019parsing, yang2020renovating, zhang2022aiparsing}. Firstly, various sperm are detected by bounding boxes. Then each sperm within a bounding box is cropped and resized to have a uniform shape through ROI Align~\cite{ren2015faster}. Finally, the sperm is segmented into individual parts. This paradigm is a top-down method~\cite{zhang2022aiparsing}. However, the top-down method is sensitive to detection errors since contexts outside a bounding box are cropped out (see Fig.~\ref{fig2}(a)). Also, the resizing step in ROI Align leads to feature distortion and inaccurate contour segmentation (see Fig.~\ref{fig2}(b)). Moreover, unlike traditional segmentation targets such as human, sperm is long and slim (high length-to-width ratio); thus, its segmentation suffers from higher context loss resulting from more detection errors and larger feature distortion because of greater aspect ratio resizing. Therefore, the top-down method produces larger segmentation errors for sperm analysis.		
	
	After segmentation, an automated sperm tail morphology measurement method is required. Sperm head parameters (e.g., length, width, and ellipticity) and midpiece parameters (e.g., length, width, and rotated angle) can be calculated through ellipse~\cite{dai2018automated} and rectangle~\cite{dai2022staining} fitting (see Fig.~\ref{fig1}(c)). In comparison, morphology measurement of sperm tail is more challenging because the tail is long, curved, with uneven in width and cannot be easily fitted. Methods have been proposed for measuring curvilinear structures~\cite{usamentiaga2012fast, li2017sub, steger1998unbiased, yang2008robust}, among which Steger-based methods~\cite{steger1998unbiased, yang2008robust} are advantageous due to their sub-pixel accuracy. In Steger-based methods, points whose first derivative reaches zero along the normal are detected as centerline points. Then, width can be calculated according to distances between centerline points and edge points along the normal, and curvature can be calculated as the change of the normal of the centerline points (see Fig.~\ref{fig1}(d)). However, endpoints detected by Steger-based methods tend to mislocate (see Fig.~\ref{fig2}(c)) because endpoints' normal is affected by gradient from the intersecting edge and not perpendicular to the line direction (see Fig.~\ref{fig2}(d)).
	
	The problems tackled in this work include: 1) how to avoid context loss and feature distortion during the instance-aware part segmentation for sperm; 2) how to accurately reconstruct endpoints for measuring sperm tail morphology parameters. To address the first challenge, a novel attention-based instance-aware part segmentation network is proposed to reconstruct lost contexts outside a bounding box and fix distorted features by refining masks generated following the ``detect-then-segment" paradigm. Specifically, the preliminary segmented masks are used to provide spatial cues for each located sperm, then merged with features extracted by a feature pyramid network (FPN) through the attention mechanism, finally refined by CNN to produce improved results. For the second challenge, an automated centerline-based tail morphology measurement method is proposed, in which an outlier filtering method and endpoint detection algorithm are designed to accurately reconstruct endpoints. 
	
	Experimental results demonstrate that the proposed network achieved $57.2\%$ AP$^{p}_{vol}$ (Average Precision based on part) on the collected dataset, outperforming the state-of-the-art top-down RP-R-CNN by $9.20\%$; and the proposed tail morphology measurement method substantially decreased measurement errors by $4.21\%, 3.02\%, 13.10\%$ for length, width and curvature, respectively compared with Steger-based methods.

	\section{RELATED WORKS}
	
	This section reviews instance-aware part segmentation works from the human parsing field, then discusses existing works on the measurement of sperm morphology parameters.
	
	\subsection{Instance-Aware Part Segmentation}
	
	The instance-aware part segmentation network was originally proposed in the human parsing field, with the aim of segmenting various human parts (e.g., arm, leg and hair) and associating each segmented part with the corresponding instance. State-of-the-art methods can be categorized into bottom-up methods and top-down methods. The bottom-up methods~\cite{gong2018instance, he2020grapy, gong2019graphonomy} first segment each part of the target as in semantic segmentation~\cite{min2021controlling,yan2022impact}, then group segmented parts into corresponding instances. For example, the part grouping network (PGN)~\cite{gong2018instance} utilizes line information to separate adjacent instances. Gray-ML~\cite{he2020grapy} and Graphonomy~\cite{gong2019graphonomy} employ graph neural networks to explore underlying label semantic relations. Although the bottom-up methods are suitable for global semantic segmentation, their instance distinction ability is poor since they lack the bounding box locating procedure.
	
	To accurately separate adjacent instances, the top-down methods first locate each instance within the image using bounding box detection, then segment parts for each located instance individually. Based on Mask R-CNN~\cite{he2017mask}, the classical instance segmentation network, Parsing R-CNN~\cite{yang2019parsing} utilizes a geometric and context encoding module for improving part segmentation accuracy. RP-R-CNN~\cite{yang2020renovating} further boosts the parsing ability by introducing a global semantic network. AIparsing~\cite{zhang2022aiparsing} utilizes FCOS~\cite{tian2020fcos}, an anchor-free detector instead of the anchor-based detector such as in Mask R-CNN to improve detection performance. Although the top-down methods have strong instance distinction ability, they suffer from context loss and feature distortion due to bounding box cropping and resizing in ROI Align.
	
	\subsection{Measurement of Sperm Morphology Parameters}
	
	Existing works mostly focused on sperm head and midpiece. In~\cite{dai2018automated, dai2022staining}, head morphology parameters and midpiece morphology parameters were calculated by ellipse and rectangle fitting. In~\cite{zhang2021quantitative}, the head-midpiece angle was calculated as the angle between the major axis of head and the major axis of midpiece. However, morphology measurement of sperm tail is challenging and mostly ignored because sperm tail is long, curved, with uneven width and cannot be easily fitted by ellipse or rectangle. The methods proposed for measuring curvilinear structures can be categorized into pixel level methods and sub-pixel level methods. Pixel level methods include skeleton extraction methods~\cite{jang2002detection} and direction template methods~\cite{haijun2003method}. To improve measurement accuracy, sub-pixel level methods were developed, including grey centroid methods~\cite{usamentiaga2012fast, li2017sub} and Steger-based methods~\cite{steger1998unbiased, yang2008robust}. The grey centroid methods first extract centroid points through vertically or horizontally line-by-line scanning, then link all centroid points to form the centerline. However, these methods perform poorly if the curvilinear structure is not directed either vertically nor horizontally~\cite{he2017robust}. In comparison, Steger-based methods calculate center points along the normal direction of centerlines, which are not affected by the direction of the curvilinear structure~\cite{xu2020line}.

	\section{METHODS}
	
	To achieve quantitative morphology measurement for sperm parts including head, midpiece and tail, two key techniques are developed in this work. A novel attention-based instance-aware part segmentation network is designed to accurately segment sperm parts; and an automated centerline-based tail morphology measurement method is developed to quantitatively calculate tail morphology parameters based on the segmented masks.
	
	\subsection{Attention-Based Instance-Aware Part Segmentation Network}
	
\begin{figure}[t]
		\centering
		\includegraphics[width=3.4in]{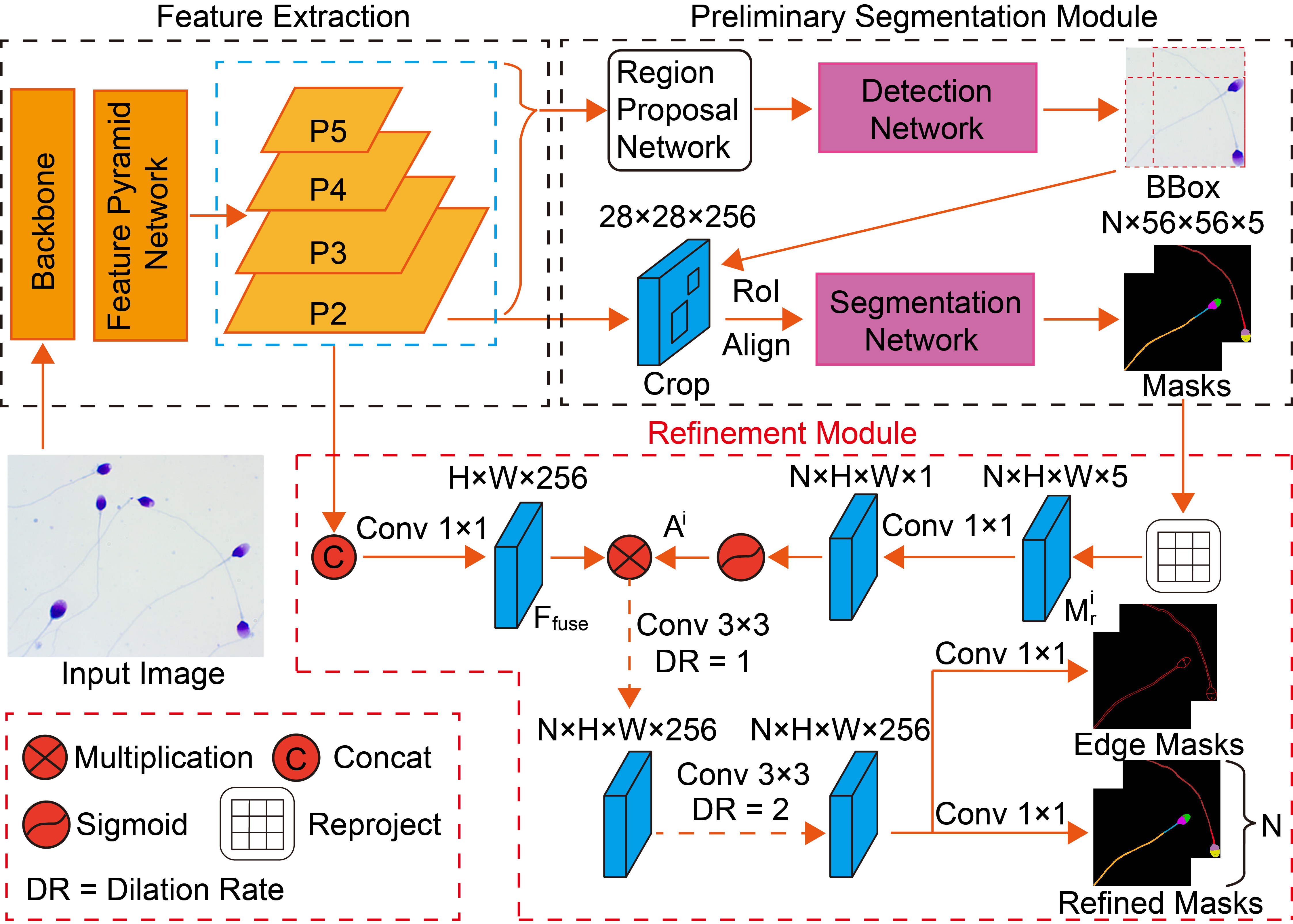}
		\caption{The structure of our proposed attention-based instance-aware part segmentation network. The convolutional backbone and FPN first extracts features from the input image then rescales extracted features to multi-scale. Next, the preliminary segmentation module generates instance-level parsing masks as in top-down methods. Finally, the refinement module refines preliminary generated masks by merging features extracted by FPN through the attention mechanism. Besides, the edge information is utilized to better separate boundary between adjacent sperm.}
		\label{fig3}
	\end{figure}

	The overall structure of the proposed network is shown in Fig.~\ref{fig3}. The input image is first fed into the convolutional backbone to extract features, then the extracted features are rescaled by the Feature Pyramid Network (FPN)~\cite{lin2017feature} to acquire multi-scale features. Next, the preliminary segmentation module generates instance-level parsing masks following the top-down methods~\cite{yang2019parsing, yang2020renovating, zhang2022aiparsing}. Firstly, sperm within the image are located by bounding boxes using candidate ROIs (region of interests) from the Region Proposal Network (RPN)~\cite{ren2015faster} in the detection branch. Secondly, each sperm within a bounding box is cropped and resized by ROI Align to have a uniform shape, finally segmented into part masks separately. It is worth noting that segmentation is performed on P2 features generated by FPN because P2 features in Fig.~\ref{fig3} have the highest resolution and contain more details. However, similar to top-down methods, the parsing masks produced by the preliminary segmentation module suffer from context loss and feature distortion (see Fig.~\ref{fig2}(a)\&(b)).
	
	To solve this issue, an attention-based refinement module is proposed to refine preliminary segmented masks. In the proposed refinement module (see Fig.~\ref{fig3}), parsing masks from the preliminary segmentation module are used to provide spatial cues of each located sperm instance, and features extracted by FPN are used to compensate for lost details. Then parsing masks and extracted features are merged through the attention mechanism and later refined by CNN to produce improved instance-aware results. Moreover, edge information is incorporated to better separate the boundary between adjacent sperm.

	Attention mechanism~\cite{hu2018squeeze, woo2018cbam} has proven to be an effective tool to selectively emphasize specific regions of an image. Therefore, the preliminary segmented masks can be used to indicate the location for each sperm during segmentation using the attention mechanism. Specifically, parsing masks $\{M^i\}_{i=1}^n$ (for $n$ sperm) are first reshaped and re-projected back to their original position in the image using the bounding box information. Then, the re-projected masks $\{M^i_{r}\}_{i=1}^n$ go through a $1 \times 1$ convolution and sigmoid activation to generate attention maps according to
	\begin{equation}
		\label{eq1}
		A^{i}=Sigmoid(Conv_{1\times1}(M^{i}_{r}))
	\end{equation}
	where $A^{i}$ denotes the attention map encoding spatial information for the $i$-th sperm. As for extracted features from FPN, multi-scale features (i.e., P2, P3, P4, P5) are merged so that targets covered by small to large receptive fields are incorporated. Specifically, P3, P4, P5 are first upsampled to the same size as P2, then P2 and upsampled P3, P4, P5 are concatenated and go through a $1 \times 1$ convolution to produce fused features according to
	\begin{equation}
		\label{eq2}
		F_{fuse} = Conv_{1\times1}(Concat[F_{P2}, F_{P3}, F_{P4}, F_{P5}])
	\end{equation}
	where $F_{fuse}$ is the fused multi-scale features. Attention maps $\{A^i\}_{i=1}^n$ are then multiplied with fused features individually to produce instance-aware part features for each sperm. Finally, the produced instance-aware part features are refined by cascaded dilated convolution layers~\cite{yu2015multi} with dilation rates of 1 and 2 to generate accurate parsing masks for each sperm by
	\begin{equation}
		\label{eq3}
		\{S^i_{j}\}_{j=1:m}=Conv_{1\times1}(DilatedConvs(A^{i} \odot F_{fuse}))
	\end{equation}
	where $\{S^i_{j}\}_{j=1:m}$ is the segmentation result for the $i$-th sperm with $m$ parts, and the final output for $n$ sperm are $\{S^i_{j}\}^{i=1:n}_{j=1:m}$. In addition, the edge-guided branch is added in parallel to the parsing branch to provide boundary information.
	
	As for the loss, weighted cross entropy loss~\cite{ronneberger2015u} and dice loss~\cite{milletari2016v} are employed since sperm have a low area ratio compared with the background. The loss for the refinement module is
	\begin{equation}
		\label{eq7}
		\mathcal{L}_{refinement}=\mathcal{L}_{parsing}^{WCE} + \mathcal{L}_{edge}^{WCE} + \lambda \mathcal{L}_{parsing}^{Dice} + \lambda \mathcal{L}_{edge}^{Dice}
	\end{equation}
	where $\mathcal{L}_{parsing}^{WCE}, \mathcal{L}_{edge}^{WCE},  \mathcal{L}_{parsing}^{Dice}, \mathcal{L}_{edge}^{Dice} $ are weighted cross entropy loss and dice loss for the parsing branch and the edge branch respectively; and $\lambda$ is a manually selected weight.
	
	\subsection{Automated Centerline-Based Tail Morphology Measurement}
	
	In Steger-based methods~\cite{steger1998unbiased, yang2008robust}, partial derivatives $r_x, r_y, r_{xx}, r_{xy}, r_{yy} $ of each pixel are first obtained by convolving the image with a discrete 2-D Gaussian partial derivative kernels. Then the normal direction $\mathbf{n}(t) = [n_x, n_y]^T$ for each pixel is determined by calculating the eigenvalues and eigenvectors of the Hessian matrix $\boldsymbol{H}(x,y) = \left[\begin{array}{cc}
		r_{xx}& r_{xy} \\
		r_{xy}& r_{yy} \\
	\end{array} \right ]$. According to~\cite{steger1998unbiased, yang2008robust}, the normal direction is perpendicular to the direction of curve-line, and centerline points are selected as points whose first derivative reaches zero (i.e., grey value reaches maximum) along normal (perpendicular to the line). Finally, points with minimum distance and minimum orientation change of normal are linked to form the centerline. However, endpoints detected by Steger-based methods tend to mislocate (see Fig.~\ref{fig2}(c)) because their normal is affected by gradient from the intersecting edge and not perpendicular to the direction of line (see Fig.~\ref{fig2}(d)).
	
	To solve this issue, an outlier filtering method and an endpoint detection algorithm are proposed. The outlier filtering method is used to filter mislocated endpoints produced by Steger-based methods, and the endpoint detection algorithm is employed to reconstruct correct endpoints.
	
\begin{figure}[t]
		\centering
		\includegraphics[width=2.8in]{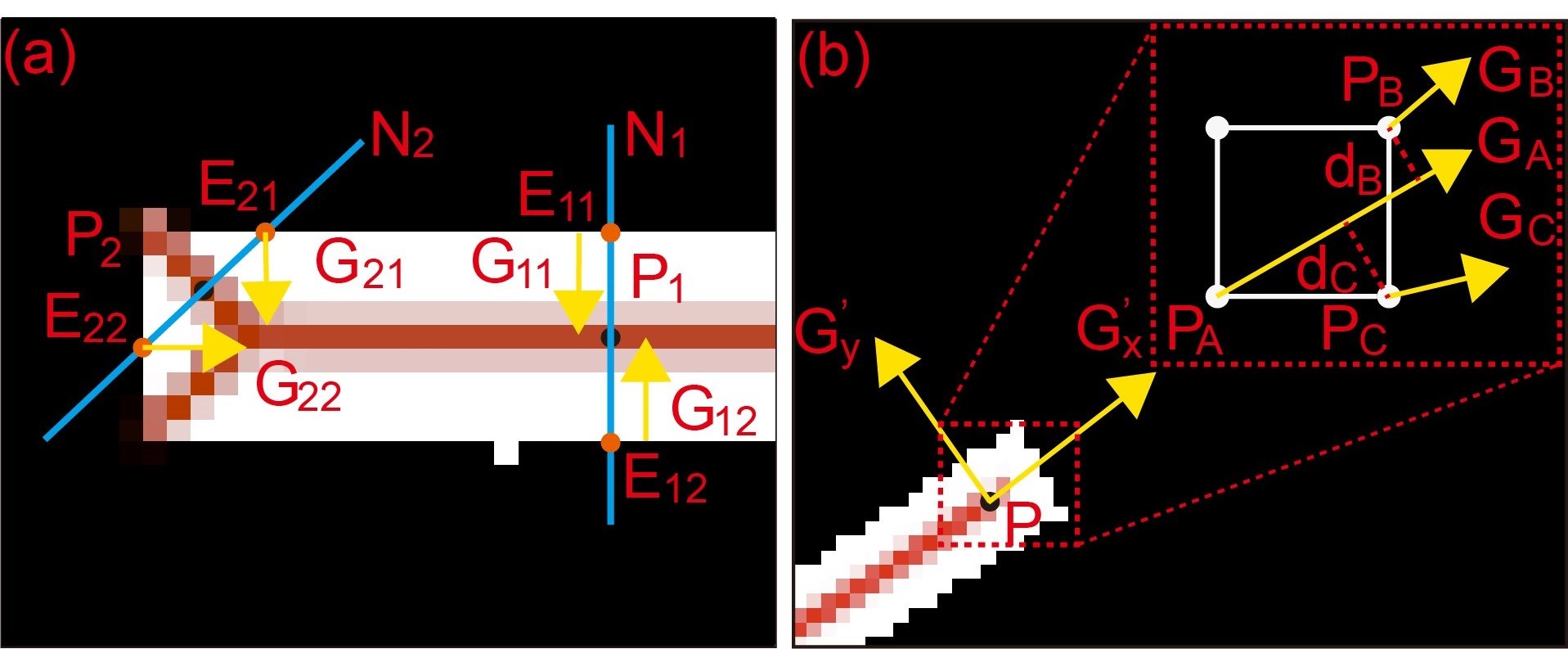}
		\caption{(a) Normal lines intersect with opposite edges ($E_{11}, E_{12}$) and intersecting edges ($E_{21}, E_{22}$) for correctly detected point $P_1$ and mislocated point $P_2$. Correspondingly, the gradient $G_{11}, G_{12}$ for $E_{11}, E_{12}$ are close to parallel, while gradient $G_{21}, G_{22}$ for $E_{21}, E_{22}$ are not. (b) The gradient for center point $P$ can be decomposed to $G^\prime_x$ and $G^\prime_y$ along the line direction and perpendicular to the line direction. According to~\cite{yang2008robust}, $G^\prime_y=0$ (center point's gradient perpendicular to the line direction is zero); therefore, the center point's gradient direction is along the direction of line.}
		\label{fig5}
	\end{figure}

	As shown in Fig.~\ref{fig5}(a), the correctly detected point lies in the middle of two opposite edges while the mislocated point lies on the diagonal of two intersecting edges. Therefore, the normal lines that cross the correctly detected and mislocated point intersect with opposite edges and intersecting edges correspondingly. Suppose $P_1$, $P_2$ are correctly detected and mislocated point, respectively, $N_1$, $N_2$ are the corresponding normal, and $E_{11}$, $E_{12}$, $E_{21}$, $E_{22}$ are corresponding intersecting points with edges. Since normal $N_1$ is perpendicular to the direction of line~\cite{steger1998unbiased}, the gradient for $E_{11}$, $E_{12}$ (on the opposite edges) are close to parallel. On the other hand, since $E_{21}$, $E_{22}$ are on intersecting edges, the gradient for $E_{21}$, $E_{22}$ are not possible to be parallel. Thus, a constraint can be formed to discriminate correctly detected and mislocated points, i.e., 
	
	\begin{figure*}[t]
		\centering
		\includegraphics[width=6.0in]{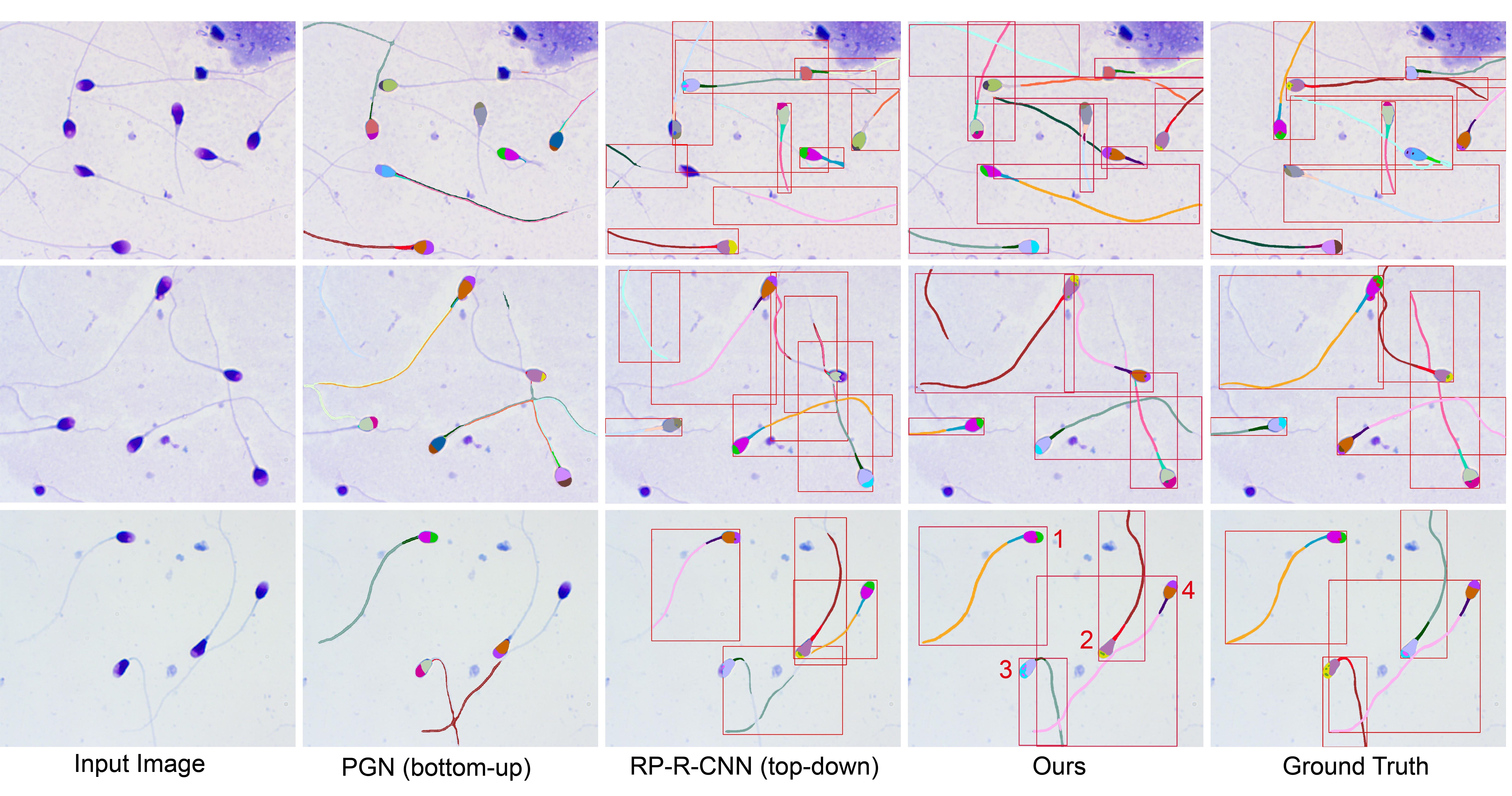}
		\caption{Qualitative comparisons of instance-aware part segmentation networks. PGN (bottom-up method) has low instance distinction and cannot separate intersecting sperm parts. RP-R-CNN (top-down method) has distorted features and contexts outside a bounding box are cropped out. In comparison, our proposed network achieves the best prediction by refining preliminary segmented masks to retrieve lost contexts outside the bounding box and fix distorted features. }
		\label{fig6}
	\end{figure*}
	\begin{equation}
		\label{eq8}
		\left\{
		\begin{aligned}
			& \cos{\alpha} = \frac{rx_1 \times rx_2 + ry_1 \times ry_2}{\sqrt{rx_1^2 + ry_1^2}\times \sqrt{rx_2^2 + ry_2^2}} \\
			& |\cos{\alpha}| \geq 0.9\\
		\end{aligned}
		\right.
	\end{equation}
	where $rx_1$, $ry_1$ and $rx_2$, $ry_2$ are first partial derivatives for edge points $E_1$ and $E_2$, receptively.
	
	After filtering mislocated points, the next step is to reconstruct missing endpoints. Without loss of generality, for any point $P$ on the centerline, its gradient $G$ can be decomposed into two directions $G^{\prime}_x$ and $G^{\prime}_y$, along the direction of the line and the direction perpendicular to the line (see Fig.~\ref{fig5}(b)). According to~\cite{steger1998unbiased, yang2008robust}, the first derivative (gradient) for center points along the normal direction (perpendicular to the line) is zero (i.e., $G^{\prime}_y = 0$), which means $G = G^{\prime}_x$ and gradient direction for point $P$ is along the direction of line. Thus, an endpoint detection algorithm is proposed by interpolating points along the direction of center point's gradient. As shown in Fig.~\ref{fig5}(b), for a point $P_A(c_x, c_y)$, if the angle of gradient is within $[0^\circ, 45^\circ]$, then point $P_B(c_x + 1, c_y + 1)$ and $P_C(c_x + 1, c_y)$ along gradient direction of $P_A$ are selected as candidate points. Then, the candidate point that minimizes $w_1 \times d + w_2 \times \beta$ is selected as the next point, where $d$ is distance between candidate point to gradient of current point, $\beta$ is the angle difference between the gradient of candidate point and current point, and $w_1, w_2$ are weights. Moreover, to avoid local minima and smooth the line, gradient for the selected next point is updated using the momentum technique~\cite{sutskever2013importance} as 
	\begin{equation}
		\label{eq9}
		g_{t+1}^{\prime} = \alpha \cdot g_t + (1 - \alpha) \cdot g_{t+1}
	\end{equation}
	where $g_t$, $g_{t+1}$ is the gradient for current point and next point, $\alpha$ is the momentum hyper-parameter, and $g_{t+1}^{\prime}$ is the updated gradient. Meanwhile, the normal for reconstructed endpoints is corrected as the direction perpendicular to gradient. This algorithm is continued until the interpolated point is out of the tail area.

	\section{EXPERIMENTS AND RESULTS}
	
	\subsection{Datasets and Evaluation Metric}
	
	In this work, we collected a new dataset\footnote{The dataset is available at https://github.com/Chenwy10/Sperm-Dataset} for the sperm part segmentation task to distinguish unique sperm instances and also segment parts (i.e., acrosome, vacuole, nucleus, midpiece and tail) for each sperm. All images were manually labelled with each part of the sperm assigned both an instance ID and a part label. Each image has the resolution of $1280 \times 1024$ and includes 5-6 sperm averagely. The dataset contains 320 images in total, with $80\%$ of them randomly selected as training samples and $20\%$ randomly selected as testing samples.
	
	\begin{table}\footnotesize \centering
		\caption{QUANTITATIVE COMPARISONS OF INSTANCE-AWARE PART SEGMENTATION NETWORKS}
		\label{table1}
		\setlength{\tabcolsep}{2pt}
		\begin{tabular}{l|l|cccc}
			\hline
			Methods&
			Backbone &
			mIOU &
			AP$^{p}_{vol}$ &
			AP$^{p}_{50}$ &
			PCP$_{50}$ \\
			\hline
			\hline
			
			Bottom-up:&&\\
			\quad PGN~\cite{gong2018instance}&
			ResNet-101
			& 57.5
			& 25.9
			& 33.9
			& 30.0
			\\
			
			\quad
			GPM~\cite{he2020grapy}&
			Xception
			& 58.2
			& -
			& -
			& -
			\\
			
			\quad
			Deeplab V3+~\cite{chen2018encoder}&
			Xception
			& 57.7
			& -
			& -
			& -
			\\
			
			\hline
			
			Top-down:&&\\
			\quad Parsing R-CNN~\cite{yang2019parsing}&
			ResNet-50
			& 54.7
			& 47.7
			& 56.7
			& 51.4
			\\
			
			\quad RP-R-CNN~\cite{yang2020renovating}&
			ResNet-50
			& 59.6
			& 48.0
			& 62.4
			& 53.9
			\\
			
			\quad AIParsing~\cite{zhang2022aiparsing}&
			ResNet-50
			& 60.7
			& 47.3
			& 57.5
			& 58.7
			\\
			
			\hline
			
			\quad \textbf{Ours} &
			ResNet-50
			& 69.1 & 57.0 & 76.1 & 68.9
			\\
			
			\quad \textbf{Ours} &
			ResNet-101& 69.3 & 57.2 & 76.8 & 69.5
			\\
			
			\hline
		\end{tabular}
		\label{tab1}
	\end{table}
	
	\begin{table*}\footnotesize \centering
		\caption{QUANTIFICATION OF MORPHOLOGY PARAMETERS OF SINGLE SPERM (AU:ARBITRARY UNIT)}
		\label{table2}
		\setlength{\tabcolsep}{2pt}
		\begin{tabular}{c|ccc|c|c|cc|c|ccc|ccc}
			\hline
			\hline
			\multirow{3}{*}{Sperm}    &
			\multicolumn{3}{c|}{Head} &
			\multicolumn{1}{c|}{Acrosome} &
			\multicolumn{1}{c|}{Nucleus} &
			\multicolumn{2}{c|}{Vacuole} &
			\multicolumn{1}{c|}{Head-Midpiece} &
			\multicolumn{3}{c|}{Midpiece} &
			\multicolumn{3}{c}{Tail}
			\\
			
			& length & width & ellipticity & area & area & number & area & angle & length & width & angle(max) & length & width & angle(max)  \\
			
			& ($\upmu$m) & ($\upmu$m) & (AU) & ($\upmu$$m^2$) & ($\upmu$$m^2$) & (AU) & ($\upmu$$m^2$) & ($^{\circ}$) & ($\upmu$m) & ($\upmu$m) & ($^{\circ}$) & ($\upmu$m) & ($\upmu$m) & ($^{\circ}$)  \\
			
			\hline
			
			1 & 4.67 & 2.69 & 1.74 & 3.54 & 6.79 & 1 & 0.33 & 27.16 & 3.68 & 0.58 & 5.27
			& 31.38 & 0.65 & 20.91 \\
			
			2 & 5.05 & 2.57 & 1.96 & 2.89 & 7.28 & 1 & 0.59 & 2.30 & 4.27 & 0.60 & 7.61
			& 27.58 & 0.59 & 18.83 \\
			
			3 & 5.08 & 2.67 & 1.90 & 4.37 & 6.48 & 2 & 0.93 & 50.60 & 2.17 & 0.43 & 37.27 & 20.85 & 0.58 & 15.83 \\
			
			4 & 4.47 & 2.61 & 1.71 & 3.57 & 6.04 & 0 & NA & 0.19 & 3.55 & 0.59 & 8.84 & 41.95 & 0.60 & 23.97 \\
			
			\hline
			\hline
			
		\end{tabular}
		\label{tab2}
	\end{table*}
	
	To evaluate the performance of our proposed network, the standard mean intersection over union (mIoU)~\cite{long2015fully} was adopted for evaluating global semantic segmentation; and the Average Precision based on part (AP$^{p}$)~\cite{zhao2018understanding} and Percentage of Correctly parsed semantic Parts (PCP)~\cite{zhao2018understanding} were used to measure instance-level segmentation accuracy. Specifically, the AP$^{p}_{50}$, AP$^{p}_{vol}$ and PCP$_{50}$  values were reported, where the first and third metric has an IoU threshold of 0.5, and the secondly metric is the mean of a series of IoU thresholds ranging from 0.1 to 0.9, with the increment of 0.1.

	\subsection{Comparisons with State-of-the-art Models}
	
	To evaluate the performance of sperm part segmentation, the proposed network was compared with several state-of-the-art instance-aware part segmentation networks, including PGN~\cite{gong2018instance}, GPM~\cite{he2020grapy}, Deeplab V3+~\cite{chen2018encoder} for bottom-up methods; and Parsing R-CNN~\cite{yang2019parsing}, RP-R-CNN~\cite{yang2020renovating}, AIParsing~\cite{zhang2022aiparsing} for top-down methods, all using our collected dataset. Qualitative and quantitative comparison results are summarized in Fig.~\ref{fig6} and Table~\ref{table1}, respectively. As can be seen, bottom-up methods gave low AP$^{p}_{vol}$, AP$^{p}_{50}$ and PCP$_{50}$ for instance-level segmentation (see Table~\ref{table1}) and cannot separate intersecting sperm parts (see Fig.~\ref{fig6}). This is because they have low instance distinction due to the lack of the bounding box locating procedure. On the other hand, top-down methods achieved significantly higher AP$^{p}_{vol}$, AP$^{p}_{50}$ and PCP$_{50}$ in instance-level segmentation compared with bottom-up methods (e.g., $22.1\%$, $28.5\%$ and $23.9\%$ gains in AP$^{p}_{vol}$, AP$^{p}_{50}$ and PCP$_{50}$ for RP-R-CNN vs. PGN). This is because top-down methods firstly locate each sperm within an image through detection and can better separate adjacent sperm.
	
	In comparison, the proposed network further outperformed top-down methods by refining preliminarily generated masks to produce improved results. Specifically, our proposed network achieved $9.7\%$, $9.2\%$, $14.4\%$ and $15.6\%$ gain in terms of mIoU, AP$^{p}_{vol}$, AP$^{p}_{50}$ and PCP$_{50}$, respectively compared with the state-of-the-art RP-R-CNN. Moreover, as shown in Fig.~\ref{fig6}, the proposed network can more effectively retrieve lost contexts outside a bounding box and fix distorted features compared with RP-R-CNN thanks to the refining procedure. Therefore, the proposed network yielded segmentation results that were closer to the ground truth with more details and less distortion.
	
	\addtolength{\topmargin}{0.07in}
	
	\subsection{Quantification of Morphology Parameters}
	
	Sperm morphology measurement was performed on the segmented masks generated from the proposed network. Quantified values for each part are summarized in Table~\ref{table2}, where Sperm 1,2,3,4 correspond to the sperm shown in the third row of Fig.~\ref{fig6}. The head and midpiece parameters were calculated through ellipse and rectangle fitting~\cite{dai2018automated, zhang2021quantitative}, while the tail parameters were calculated using the proposed automated centerline-based tail morphology measurement method (see demo in Video).
	
\begin{figure}[t]
		\centering
		\includegraphics[width=3.4in]{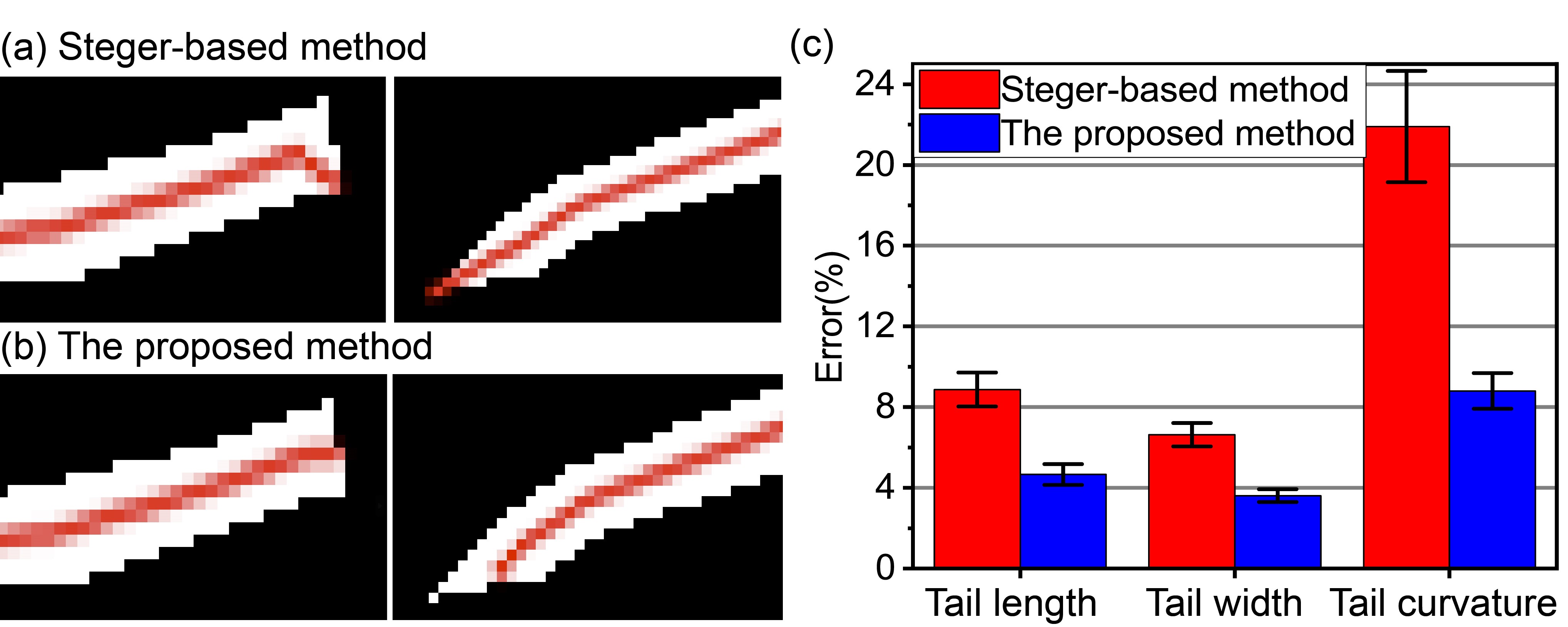}
		\caption{(a) \& (b) Qualitative comparison results for Steger-based methods and our proposed method. The proposed method can effectively fix mislocated endpoints. (c) Quantitative comparison of errors for measuring tail length, width and curvature with Steger-based methods and the proposed method. The experiment was conducted on 50 sperm with manual benchmarking.}
		\label{fig7}
	\end{figure}

	To evaluate the performance of tail morphology measurement, an additional experiment was conducted on 50 sperm using classical Steger-based methods and our proposed method. In this experiment, the length, width and curvature of sperm tail were used for quantitative comparison, with the benchmark data obtained by manually benchmarking zoomed-in masks in ImageJ with best care~\cite{dai2018automated}. Qualitative and quantitative comparison results are summarized in Fig.~\ref{fig7}(a)(b)(c). The errors of tail length, width and curvature measurements with the Steger-based method were $8.87\%, 6.63\%, 21.90\%$, respectively. These error were mostly caused by mislocated endpoints and had significant impact on curvature measurement (see Fig.~\ref{fig7}(a)). In comparison, our proposed method successfully reconstructed mislocated endpoints using the proposed outlier filtering method and endpoint detection algorithm (see Fig.~\ref{fig7}(b)), and substantially decreased measurement errors by $4.21\%, 3.02\%, 13.10\%$ for tail length, width and curvature, respectively (see Fig.~\ref{fig7}(c)). The errors for tail measurement with our proposed method were $<10\%$ for length, width and curvature, well meeting the requirement for determining normal/abnormal  sperm~\cite{dai2018automated}.
	
	\section{CONCLUSION}
	
	This paper presented automated techniques for quantitatively measuring morphology parameters for each part of a sperm. An attention-based instance-aware part segmentation network was proposed to accurately segment each part of the sperm by avoiding context loss and feature distortion. An automated centerline-based tail morphology measurement method was designed to quantitatively calculate tail parameters based on the segmented masks. Experiments showed  that the proposed network outperformed existing state-of-the-art networks, and the proposed automated centerline-based tail morphology measurement method was capable of accurately quantifying tail morphology parameters. 
	
	\addtolength{\textheight}{-11cm}   
	



	\bibliographystyle{IEEEtran}
	\bibliography{mybibfile}

\end{document}